\documentclass[10pt,twocolumn,letterpaper]{article}
\usepackage{wacv}
\usepackage{times}
\usepackage{epsfig}
\usepackage{graphicx}
\usepackage{subfigure}
\usepackage{amsmath}
\usepackage{amssymb}
\usepackage{booktabs}
\usepackage{wrapfig}
\usepackage[accsupp]{axessibility}
\def\shortname{FAN-Trans}

\def\one{(\uppercase\expandafter{\romannumeral1})}
\def\two{(\uppercase\expandafter{\romannumeral2})}
\def\three{(\uppercase\expandafter{\romannumeral3})}

\def\Vo{C\_T1\_$C_{o2o}$\_T2\_$C_{o2o}$}  \def\Vm{C\_T1\_$C_{o2m}$\_T2\_$C_{o2m}$}  \def\Vom{C\_T1\_$C_{o2o}$\_T2\_$C_{o2m}$}

\makeatletter
\def\blfootnote{\xdef\@thefnmark{}\@footnotetext}
\DeclareRobustCommand\onedot{\futurelet\@let@token\@onedot}
\def\@onedot{\ifx\@let@token.\else.\null\fi\xspace}
 
\def\ie{{\em i.e}\onedot}

\def\etal{{\em et al}\onedot}

\def\wacvPaperID{926} 

\wacvalgorithmstrack   

\wacvfinalcopy 

\ifwacvfinal
\usepackage[breaklinks=true,bookmarks=false]{hyperref}
\else
\usepackage[pagebackref=true,breaklinks=true,colorlinks,bookmarks=false]{hyperref}
\fi

\begin{document}

\title{\shortname{}: Online Knowledge Distillation for Facial Action Unit Detection}

\author{
Jing Yang \textsuperscript{1}\\ 
\tt\small y.jing2016@gmail.com
\and
Jie Shen \textsuperscript{2}\\
{\tt\small jie.shen07@imperial.ac.uk}
\and
Yiming Lin \textsuperscript{2}\\
{\tt\small yl1915@ic.ac.uk}
\and
Yordan Hristov \\
{\tt\small yshristov@gmail.com}
\and
Maja Pantic\textsuperscript{2}\\
{\tt\small maja.pantic@gmail.com}\\
\vspace{-5mm}
\and
\textsuperscript{1}University of Nottingham, UK \hspace{5mm} \textsuperscript{2}Imperial College London, UK \\
}

\maketitle
\thispagestyle{empty}

\begin{abstract}
\blfootnote{$^{\dagger}$ Corresponding author.}
Due to its importance in facial behaviour analysis, facial action unit (AU) detection has attracted increasing attention from the research community. 
Leveraging the online knowledge distillation framework, we propose the ``\shortname{}" method for AU detection. 
Our model consists of a hybrid network of convolution and transformer blocks to learn per-AU features and to model AU co-occurrences. 
The model uses a pre-trained face alignment network as the feature extractor. 
After further transformation by a small learnable add-on convolutional subnet, the per-AU features are fed into transformer blocks to enhance their representation. 
As multiple AUs often appear together, we propose a learnable attention drop mechanism in the transformer block to learn the correlation between the features for different AUs.
We also design a classifier that predicts AU presence by considering all AUs' features, to explicitly capture label dependencies. 
Finally, we make the attempt of adapting online knowledge distillation in the training stage for this task, further improving the model's performance. 
Experiments on the BP4D and DISFA datasets have demonstrated the effectiveness of proposed method. 

\end{abstract}
\vspace{-5mm}
\section{Introduction}
Facial behaviour is a natural and effective way to convey emotions, sentiments, and mental states in face-to-face communications. 
Due to its great potential in human-robot interaction, digital marketing, and psychological and behavioural research, 
automatic facial behaviour analysis has attracted increasing attention from both the academic community and the industry. 
Among different representations, facial action units (AU) provide the most comprehensive, expressive and objective descriptors for facial behaviours.
They are defined over muscle movements according to the anatomy of human faces.
As such, a robust AU detection method is important in facial behaviour understanding.

\begin{figure}[t]
  \centering
  \includegraphics[width=0.98\linewidth]{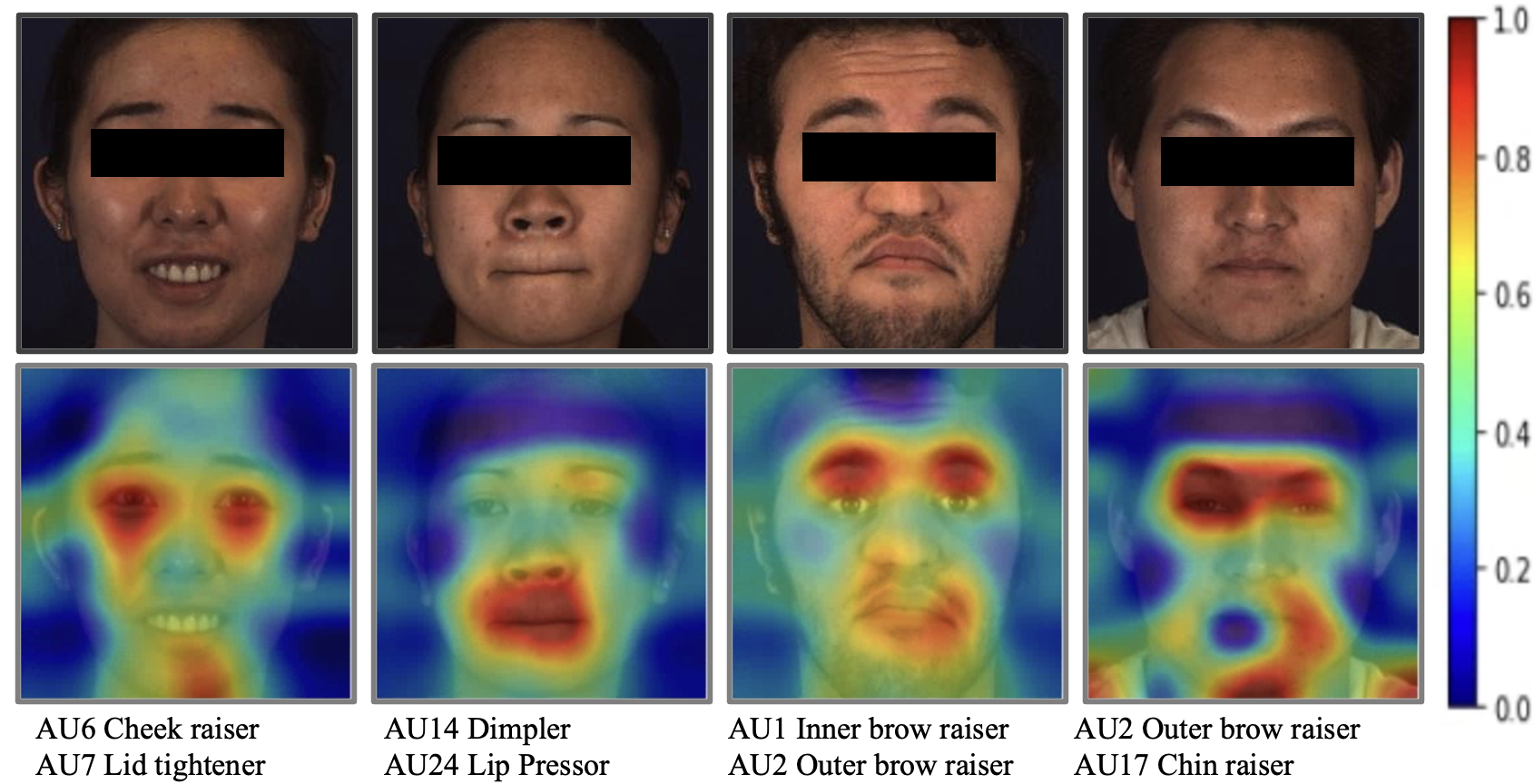}
  \vspace*{-2pt}
  \caption{Visualization of attention maps of \shortname{} via Score-CAM \cite{score-cam}. The result is achieved by placing the target layer on resolution $4 \times 4$. From top to bottom, each row denotes an input image, its spatial attention and activated action units, in respect. \shortname{} shows a potential high concentration at the facial component that indicates its ability to learn where to focus without using explicit attention modules \cite{TransAU} or manual region allocations \cite{PIAP,SRERL}. 
  For example, AU1 and AU2 emphasize the eyebrow, AU7 is around the eye while AU24 highlight the mouth.}
  \vspace{-6mm}
  \label{fig:attention}
\end{figure}

In Facial Action Coding System (FACS)~\cite{FACS}, AUs are determined by appearance changes (\ie geometry shape,
textures) caused by facial muscle movements on a human face. 
Those changes are subtle, local and connected.
For example, due to the underlying facial anatomy, AU1 and AU2 usually appear together because they are controlled by same muscle.
In order to align with this area, the network design is desired to capture local AU features and consider the property of AU co-occurrences.

Facial landmarks, representing semantic key-points on a human face, are recognized as AU active locations.
They are often used to crop the region of interest (ROI) for AUs \cite{DRML,zhang2018classifier,DSIN} to reduce distraction from unrelated facial areas.
Nevertheless, even after aligning faces to a common reference frame, precise AU localisation remains a challenge due to head pose and view variations, which will adversely affecting cropping ROI for AUs.
Thus, we propose to extract intermediate representations from a pre-trained facial landmark detector for AU features learning.
Due to the nature of face alignment task, these features are face specific and landmark-focused.


To consider the AU co-occurrences, previous works often used a standalone module to explicitly model the label or feature correlation between different AUs \cite{wang2013capturing,walecki2017deep,SRERL}. 
Recently, motivated by transformer network's effectiveness in learning the correlations between distant patches in the image classification task \cite{dosovitskiy2020image,vaswani2017attention},
TransAU \cite{TransAU} and TAM \cite{song2021facial} pioneered in applying it to facial AU detection. 
Built upon this idea, we further introduce a learnable binary attention module to the transformer block to enhance its capability.
This leads to more discriminative AU features.

Seminal works \cite{DRML,DSIN,JAANet,JAANet+,SRERL,PIAP,TransAU} are dedicated to architecture or loss designs to tackle the aforementioned issues, but to the best of our knowledge, none has tried to improve the learning procedure using online knowledge distillation (OKD). 
For AU detection, no high capacity model is readily available because increasing the number of parameters does not necessarily lead to higher accuracy due to the over-fitting problem caused by the limited amount of training data \cite{walecki2017deep}. 
To address this, we propose to use OKD \cite{zhang2019your,zhang2019scan,lan2018knowledge,li2021online} in our method.
In contrast to two-stage knowledge distillation \cite{KD}, OKD can boost the model's accuracy without requiring a pre-trained teacher. 
It has achieved impressive results in tasks like object classification \cite{lan2018knowledge,chen2020online}, human pose estimation \cite{li2021online} but it has not been explored in the context of facial AU detection.

To this end, we formulate the AU detection task as a multi-label classification problem within the online knowledge distillation framework. 
Using a pre-trained face alignment network as the feature extractor, we only add a small subnet to learn per-AU features, as the intermediate feature maps produced by the face alignment network already provide rich shape and contextual information.
To model AU co-occurrences, we propose a learnable attention drop mechanism on the self-attention module in the transformer block and significantly enhances the model's performance by decreasing homogeneity among AU features.
Additionally, we analyse two classifiers with one to predict per AU's activation on a single AU's feature and the other predicts all AUs' existence based on it.
We show that the latter achieves superior performance as it also learns the co-occurrences of AUs in the label space. 
Last but not least, we apply the OKD framework with diverse classifiers in the training stage to further improve the model's accuracy, without incurring additional cost at inference time.

Our method, coined \shortname{}, has achieved a new state-of-the-art performance on public benchmarks. 
The visualization of attention maps learned by \shortname{} is shown in Figure \ref{fig:attention}. 
Without explicitly ROI assigned for different AUs \cite{TransAU,PIAP,SRERL}, FAN-Trans can learn where to focus based on only supervision from AU labels. 


The main \textbf{contributions} of this paper are listed below.
\begin{enumerate}
\vspace{-2mm}
\item To our best knowledge, we are the \textit{first} to adopt OKD with diverse peers designed for AU detection, improving its performance via ensemble learning.
\vspace{-2mm}
\item Instead of manually assigning pre-defined regions to different AUs, \shortname{} builds upon a pretrained multi-scale face alignment features to automatically learn the spatial correspondence between AUs and the underlying facial parts.
\vspace{-2mm}
\item We propose to exploit both feature and label correlations in solving the AU detection task. 
For features, we design a transformer block with a learnable binary mask to learn sub connections of AUs; For labels, we devise a classifier to predict single AU's activation on features of all AUs.
\vspace{-2mm}
\item Through extensive experiments, we demonstrate the effectiveness of our proposed method on two widely used benchmark datasets: BP4D~\cite{BP4D} and DISFA~\cite{DISFA}.
\end{enumerate}

\section{Related works}
\noindent
\textbf{Regional feature representation}
Since AUs are defined in FACS \cite{FACS} as muscular activation on the face, the AU detection task can be formulated as a classification task on local features extracted around facial landmarks. 
The early works utilised handcraft features. A typical pipeline in \cite{zhang2018classifier} was to first align crop faces, then extract handcraft features in a predefined patch around landmarks, and finally enhance feature representation by fusing texture features with geometry features formed by landmark coordinates.  
Recently, deep models have been widely used to capture local appearance changes for AU detection. 
For example, Zhao \etal \cite{DRML} designed a region layer to induce specific facial regions for identifying different AUs. 
Shao \etal \cite{JAANet+} developed an end-to-end multi-task framework to jointly do AU detection and face alignment, and the heat-maps were used to pre-define the ROI which was further refined in model optimization.
The aforementioned works linked AU detection with the ROI features, which have proven to be notably effective for AU detection. 
Building upon this insight, we propose to extract informative face shape and contextual priors from a pre-trained face alignment model and let the network automatically assign AU features during optimization.

\begin{figure*}[t]
\centering
\includegraphics[height=0.45\textwidth]{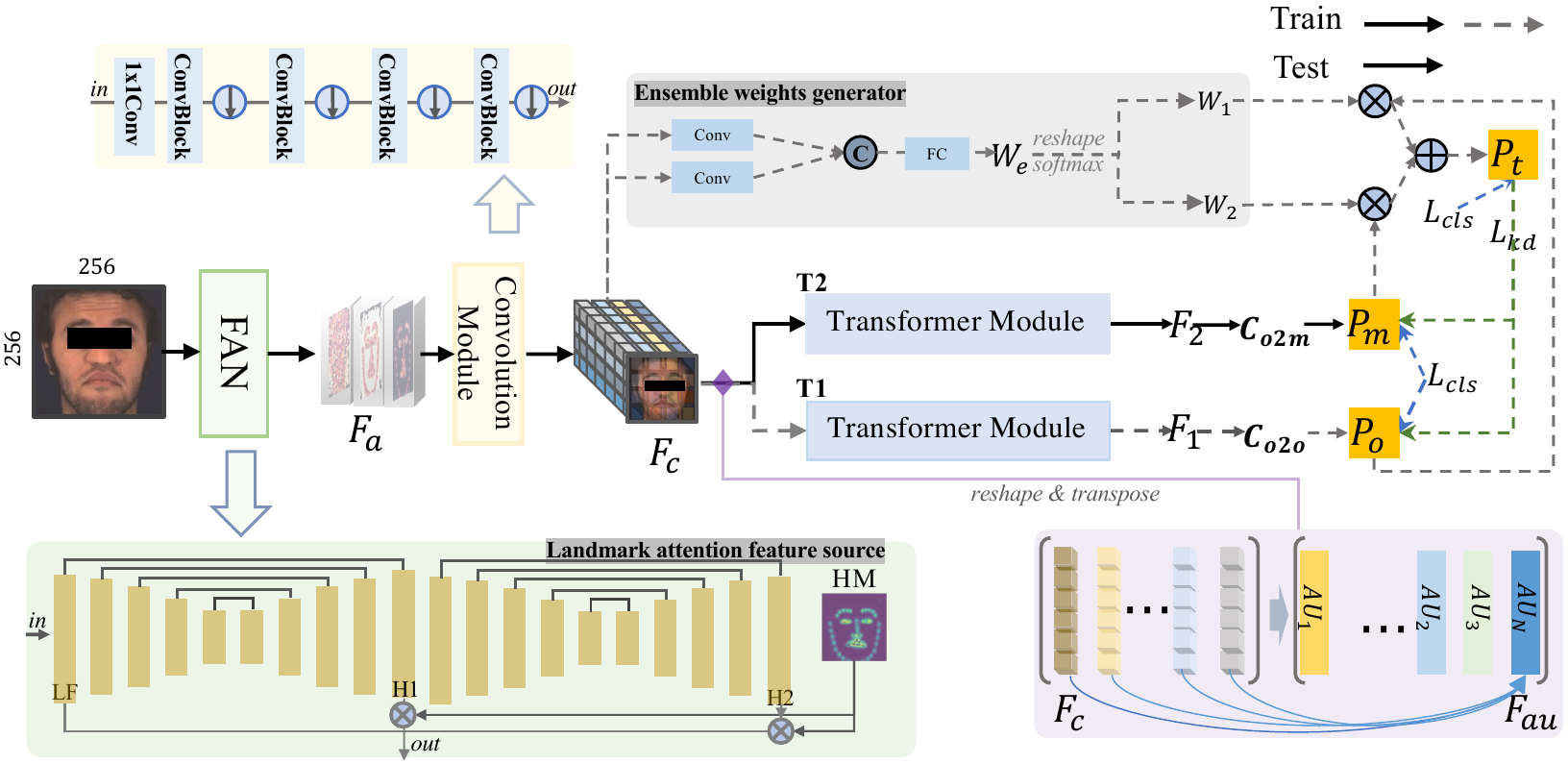}
\caption{Overview of \shortname{}.
In the training stage, firstly, a face image is fed into a pre-trained face alignment model to extract feature maps $\mathbf{F}_{a}$. 
Secondly, $\mathbf{F}_{a}$ is passed through a convolution module to learn compact feature representation $\mathbf{F}_{c}$. 
Thirdly, $\mathbf{F}_{c}$ is reshaped and transposed to a sequence of vectors, and is further projected into a set of AU features $\mathbf{F}_{au}$ by a linear transformation. 
$\mathbf{F}_{au}$ is fed into two transformer branches with diverse classifiers $C_{o2o}$ and $C_{o2m}$ to get predictions $\mathbf{P}_{o}$ and $\mathbf{P}_{m}$. 
The ensemble target $\mathbf{P}_{t}$ is a weighted ($W_{1}$, $W_{2}$) combination of $\mathbf{P}_{o}$ and $\mathbf{P}_{m}$. 
Finally, all learnable parameters are optimized by three classification losses $\mathcal{L}_{cls}$ and two knowledge distillation losses $\mathcal{L}_{kd}$. 
Note that only a single pathway remains at inference time. All auxiliaries modules enclosed in dashed lines are discarded after training.
By doing so, regional feature learning, AU-occurrences modeling, efficient training technique are integrated into one end-to-end trainable pipeline.}
\vspace*{-10pt}
\label{fig:overview}
\end{figure*}

\noindent
\textbf{AU co-occurrences modeling}
Due to the underlying facial anatomy, activations of different AUs are often correlated. 
Therefore, instead of detecting each AU independently, learning the co-occurrences of AUs can be incorporated either into the the network design, or as a standalone refinement step.
Some works~\cite{wang2013capturing,walecki2017deep} realize it by attaching an explicit module on initial predictions with probabilistic graphical models.
For example, the early attempt~\cite{wang2013capturing} exploited the restricted Boltzmann machine to learn AU relations. Similarly, \cite{walecki2017deep} appended conditional random field on the top of fully connected layer output to force AU dependencies. 
Others \cite{SRERL,TransAU} concentrate on learning semantic correlation among AUs. 
For example, SRERL \cite{SRERL} fed AU features to Gated Graph Neural Network built on defined knowledge-graph in an offline manner. 
More recently, TransAU \cite{TransAU} introduced transformer \cite{dosovitskiy2020image} into the facial AU detection task due to its particular efficacy in capturing dependencies among distant patches. 
Our method uses the transformer but we go a step further by discriminating AU features with a learnable drop attention in self-attention module. 
Besides, our transformer network is trained with the automatically learned AU features instead of cropped features from pre-defined ROIs. 

\noindent
\textbf{Online knowledge distillation}
Different from vanilla KD \cite{KD}, which uses a pre-trained high capacity teacher network to guide the learning of a low-capacity student, there is no explicit teacher network in OKD.
One category of OKD is to train with the ensemble output of the students sharing similar network configurations. 
For example, ONE \cite{lan2018knowledge} was the pioneer who constructed a single multi-branch network to let each branch learn from the ensemble distribution in the classification task. 
This idea was extended in the same or heterogeneous settings and applied in the human pose estimation task \cite{li2021online}. 
Our model is thus an instantiation of OKD in the specific context of AU recognition. We further maximize the capacity of OKD by increasing the divergence of peers.

\section{\shortname{}}
\label{sec:fantrans}
The architecture of the proposed \shortname{} is shown in Figure \ref{fig:overview}. 
This section will describe each component and the rational behind in detail.

\subsection{Landmark Attention Feature Extraction}
\label{sec:lafe}
Facial landmarks represent the semantically salient regions of a human face. 
Given that AUs are present in local regions around facial landmarks, previous approaches~\cite{DRML,SRERL} used landmarks to predefine ROI for AUs.
Three drawbacks could be observed in such assignment: extra time cost, sensitivity to precision of landmarks, and difficulty to embrace unregistered AUs.
To avoid these, we take intermediate features from a pre-trained facial landmark model and learn AU embeddings on them. 

Concretely, we leverage the stacked-hourglass-based FAN \cite{hourglass,FAN} model for this purpose, which has been explored as the feature extractor for face recognition \cite{FANFace} and face emotion recognition \cite{emonet}.
FAN was trained for landmarks localization with heat-maps (Gaussian circle peaking at keypoints) as supervision on a large corpus of facial data, covering the full range of poses. 
Unlike models trained on ImageNet with a classification task which are generic (\ie not specific to faces) and coarse, 
FAN features 
\textbf{(1)} capture the finer grained aspects of the face --- a direct consequence of the face alignment task;  
\textbf{(2)} are robust to appearance variations from pose, illumination, color as the model was pre-trained on a large variety of facial poses;
\textbf{(3)} closely align with AU detection --- having a good localization of the AU region correlates with higher AU accuracy.


As illustrated in Figure \ref{fig:overview}, we obtain source of AU features by combining intermediate features (LF, H1, H2) and heat-maps (HM) from the pre-trained FAN.
Specially, we first aggregate heat-maps to a single plane, then multiply it with penultimate layer's outputs (H1,H2), and finally concatenate these high level features with the low-level features (LF) to obtain $\mathbf{F}_{a} \in \mathbb{R}^{D_{a} \times H_{a} \times W_{a}}$. 
We use ``a" as a subscript to represent the features from FAN. 
$\mathbf{F}, D, H, W$ describe the feature tensor, and its channel, height, width respectively. 
We drop the symbol of batch size for brevity. 

\subsection{A Hybrid Network of Convolution and Transformer modules}
\label{sec:hybrid}
Based on the AU definition, a successful AU detector is supposed to learn features around AU active regions and consider the property of AU co-occurrences.
we use a hybrid convolution + transformer structure for learning the AU features, in which convolution is to obtain abstract face representations from $\mathbf{F}_{a}$ and transformer \cite{yang2021transpose,lin2021end} is to learn AU co-occurrences in feature space.

As illustrated in Figure \ref{fig:overview}, the convolution module is composed of several convolution layers: one 1x1 convolution layer to decrease the channel dimension from $D_{a}$ to $0.25D_{a}$; four convolution layers followed by a max-pooling to reduce spatial size by 4 times.
The advantages of this module are from two aspects: 
First, it makes the feature representations more abstract by enlarging the receptive field (the resolution decreases from $64\times64$ to $4\times4$);
Second, it reduces the computational complexity by decreasing feature dimension as self-attention in transformer operates across sequential tokens by drawing pairwise interactions. 
Given a input tensor $\mathbf{F}_{a}$, the output of convolution module is $\mathbf{F}_{c} \in \mathbb{R}^{D_{c} \times H_{c} \times W_{c}}$. 

Furthermore, we implicitly allocate per AU embedding with a linear transformation. 
We first flatten 3D $\mathbf{F}_{c}$ across spatial dimension to 2D tensor $\mathbf{F}_{c} \in \mathbb{R}^{D_{c}\times H_{c}W_{c}}$, then apply a linear layer on the 2rd dimension to transform $H_{c}W_{c}$ to number of AUs $N_{au}$.
This is formulated by:
\vspace{-2mm}
\begin{equation}
    \mathbf{F}_{au} = \mathbf{F}_{c} \times \mathbf{W}_{au},
\end{equation}
where $\mathbf{W}_{au} \in \mathbb{R}^{H_{c}W_{c} \times N_{au}}$, and $\mathbf{F}_{au} \in \mathbb{R}^{D_{c} \times N_{au}}$. 
By doing so, per AU feature is associated with $\mathbf{F}_{c}$ with learnable parameters optimized by the task loss. 
Every vector $D_{c} \times 1$ is considered as one AU's representation.

\begin{figure}[t]
  \centering
  \includegraphics[width=0.8\linewidth]{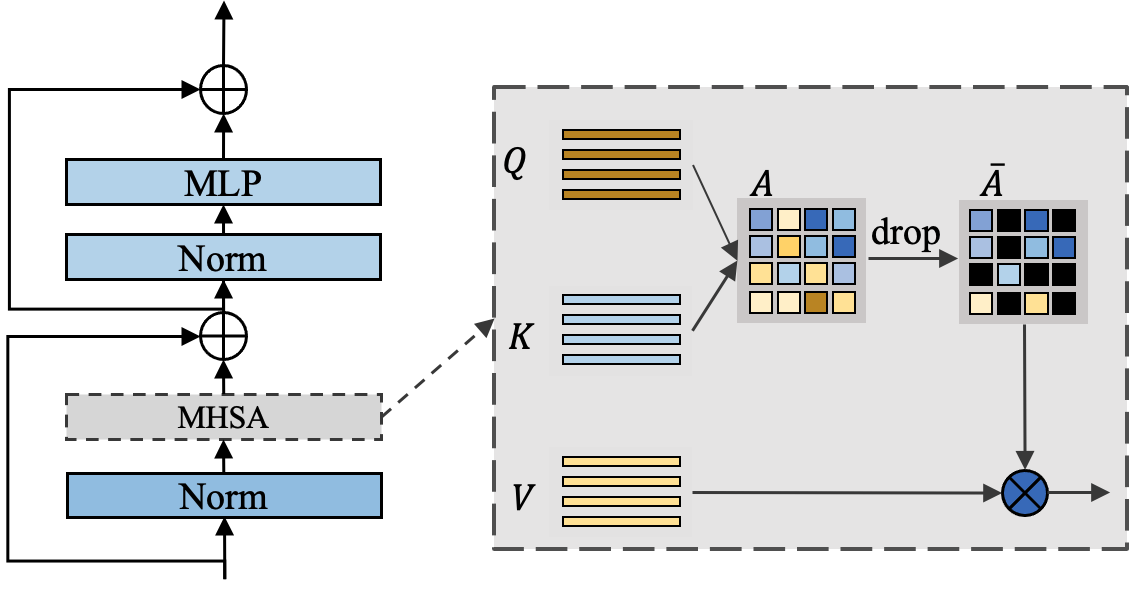}
  \caption{Encode block (transblock) in transformer module. We assume $N_{au}=4$. For a 4x4 attention map $\mathbf{A}$, we explore different attention drop mechanisms on $\mathbf{A}$ to find the best choice $\mathbf{\overline{A}}$ for facial AU detection task.}
  \vspace*{-10pt}
  \label{fig:tranformer}
\end{figure}

The transformer module is exploited here to model the AU co-occurrences. 
A learnable position embedding is applied on $\mathbf{F}_c$.
The main component of the transformer \cite{dosovitskiy2020image} module is the stacked encode blocks (transblock) composed of multi-head self-attention (MHSA) and multi-layer projection (MLP) (see Figure \ref{fig:tranformer}).
To utilize it, we apply transpose operation on $\mathbf{F}_{au} \in \mathbb{R}^{D_{c} \times N_{au}}$ to obtain $\mathbf{F}_{au} \in \mathbb{R}^{\in N_{au} \times D_{c}}$ as transformer here is to modeling connections among AU features. 
For MHSA module, linear layers are applied on $\mathbf{F}_{au}$ to obtain queries $\mathbf{Q} \in \mathbb{R}^{N_{au} \times D_{q}}$, keys $\mathbf{K} \in \mathbb{R}^{N_{au} \times D_{k}}$ and values $\mathbf{V} \in \mathbb{R}^{N_{au} \times D_{v}}$ ($D_{q}==D_{k}$), then the attention map $\mathbf{A}$ is calculated through:
\begin{equation}
    \mathbf{A} = \operatorname{softmax}(\frac{\mathbf{Q}\mathbf{K}^\top}{\sqrt{D_{q}}}),
\label{eq:A}
\end{equation}
where $\mathbf{A} \in \mathbb{R} ^{N_{au} \times N_{au}}$ with $\mathbf{A}[i,:]$ representing $\mathbf{Q}[i,:]$'s correlation with $\mathbf{K}$, and $\mathbf{A}[:,i]$ representation $\mathbf{Q}$'s correlation with $\mathbf{K}[i,:]$. 
We further propose a learnable attention drop mechanism by multiplying $\mathbf{A}$ with learned binary mask $\mathbf{M}\in \mathbb{R} ^{N_{au} \times 1}$ to drop some AUs connections with others.
\begin{equation}
    \mathbf{\overline{A}} = \mathbf{M} * \mathbf{A}.
\end{equation}
where $\mathbf{\overline{A}}_{i,j}=\mathbf{M}_{j}\mathbf{A}_{i,j}$.
This is supposed to increase the difference among AU features:

\subsection{Online Knowledge Distillation with Diverse Classifiers}
\label{sec:okd}
The main structure is a two-branch architecture, which is illustrated in Figure \ref{fig:overview}.
We feed $\mathbf{F}_{au}$ into two diverse peers ($\mathbf{T1}$ and $\mathbf{T2}$).
We denote the outputs of $\mathbf{T1}$ and $\mathbf{T2}$ as $\mathbf{F}_{1} \in \mathbb{R}^{N_{au} \times D_{t}}$ and $\mathbf{F}_{2} \in \mathbb{R}^{ N_{au} \times D_{t}}$, respectively. 
We design the diverse classifiers as follows:

\noindent
\textbf{One-to-One classifier ($C_{o2o}$):} 
We pass $\mathbf{F}_{1}$ through a fully connected layer $\mathbf{W}_{o} \in \mathbb{R}^{D_{t} \times 1}$, and the prediction is computed as
\begin{equation}
    \mathbf{P}_{o} = \mathbf{F}_{1} \times \mathbf{W}_{o},
\end{equation}
where $\mathbf{P}_{o} \in \mathbb{R}^{N_{au} \times 1}$ with i-th element  $\mathbf{F}_{1}[i]$ predicting i-th AU's activation.

\begin{figure}[h]
  \centering
  \includegraphics[width=0.8\linewidth]{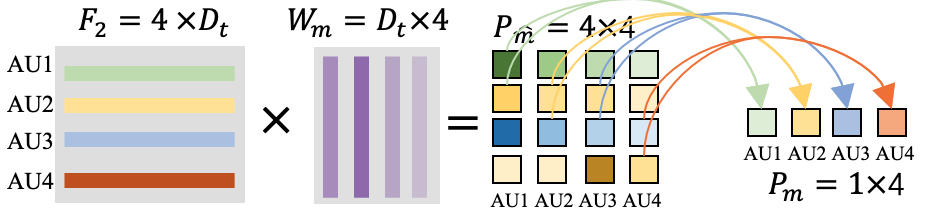}
   \caption{Steps for calculating $\mathbf{P}_{m}$. We assume $N_{au}=4$ for illustration purposes. $\mathbf{F}_{2}[i,:]$ and $\mathbf{W}_{m}[:,i]$ denote i-th AU's feature and i-th AU's prototype, respectively. Hence, i-th AU's feature ($\mathbf{F}_{2}[i,:]$) contributes to the logits of all AUs ($\mathbf{P}_{\hat{m}}[i,:]$) and all AUs' features contribute to the logits of j-th AU ($\mathbf{P}_{\hat{m}}[:,j]$). To avoid noise accumulation, $\mathbf{P}_{m}[i]$ is aggregated (summed) from the highest k (k=2) values in $\mathbf{P}_{\hat{m}}[:,i]$.}
   \label{fig:pm}
\end{figure}

\noindent
\textbf{One-to-Many classifier ($C_{o2m}$):} We multiply $\mathbf{F}_{2}$ with a transformation matrix $\mathbf{W}_{m} \in \mathbb{R}^{D_{t} \times N_{au}}$, and the prediction is calculated as
\begin{equation}
    \mathbf{P}_{m} = \operatorname{TopK}(\mathbf{F}_{2} \times \mathbf{W}_{m}),
\end{equation}
where $\mathbf{P}_{m} \in \mathbb{R}^{1 \times N_{au}}$. 
The output of $\mathbf{F}_{2} \times \mathbf{W}_{m}$ is $\mathbf{P}_{\hat{m}} \in \mathbb{R}^{N_{au} \times N_{au}}$ and \rm {TopK} is a selection function.
The i-th slice of $\mathbf{F}_{2}[i,:]$ predicts the existence of all AUs. $\mathbf{P}_{\hat{m}}[:,i]$ denotes the i-th AU's evidence from all AU features.
\rm {TopK} operation summarizes the highest k confidence in $\mathbf{P}_{\hat{m}}[:,i]$ to get the final prediction for i-th AU:
\begin{equation}
    \mathbf{P}_{m}[i] = \sum_{j=0}^{K_{i}} \mathbf{P}_{\hat{m}}[:,j],
\end{equation}
where $K_{i}$ denotes the top-k index per $i$-th column.
The illustration of $\mathbf{P}_{m}$'s calculation is in Figure \ref{fig:pm}. 

\vspace{-2mm}
\noindent
\textbf{Online knowledge distillation}
Based on the two-branch formulation, we have two transformer modules with the same number of transblocks but different AU classifiers. 
Both branches share the convolution module for compact feature extraction and per AU feature assignment. 
Inspired by previous works \cite{lan2018knowledge,li2021online}, the ensemble weight generator is placed on $\mathbf{F}_{c}$. 
Its structure is depicted in Figure \ref{fig:overview} (Ensemble weights generator). 
Specifically, it contains two splits to capture features across different receptive fields, concatenation operation to enrich representation, and fully connected layer to produce weights $\mathbf{W}_{e} \in \mathbb{R}^{2 \times N_{au}}$. Afterwards, the Softmax is used to normalize the weights. 
Finally, the ensemble weights $\mathbf{W}_{e}$ are split into $\mathbf{W}_{1} \in \mathbb{R}^{1 \times N_{au}}$ and $\mathbf{W}_{2} \in \mathbb{R}^{1 \times N_{au}}$ and the ensemble target is computed by element-wise summation:
\begin{equation}
    \mathbf{P}_{t}= \mathbf{W}_{1}*\mathbf{P}_{o}+\mathbf{W}_{2}*\mathbf{P}_{m}^\top.
\end{equation}
The OKD is formulated as the Kullback-Leibler (KL) divergence loss between $\mathbf{P}_{t}$ and the student's prediction ($\mathbf{P}_{*} \subseteq \left \{ \mathbf{P}_{o},\mathbf{P}_{m} \right \}$):
\begin{equation}
    \mathcal{L}_{KD} = \operatorname{KL}(\mathbf{P}_{*},\mathbf{P}_{t}).
\end{equation}
The multi-label classification loss is formulated as weighted binary cross entropy loss with ground-truth label:
\begin{equation}
    \mathcal{L}_{cls}= -\sum_{i}^{N_{au}}w_{i}[y_{i}\operatorname{log} p_{i}+(1-y_{i})\operatorname{log}(1-p_{i}))],
\end{equation}
where $p_{i}$ denotes i-th AU's prediction, which has three sources $\mathbf{P}_{o}$, $\mathbf{P}_{m}$, and $\mathbf{P}_{t}$. 
$y_{i}$ is the ground-truth label for i-th AU (1 denotes AU appears and 0 denotes AU does not present). $w_{i}$ is the class weight for each AU based on AU's occurrence to balance training \cite{JAANet}.

The final network is trained in an end-to-end manner by minimizing the following cost function:
\begin{equation}
    \mathcal{L}_{total} = \mathcal{L}_{cls}+\lambda\mathcal{L}_{kd},
\end{equation}
in which $\lambda$ is hyper-parameter to balance $\mathcal{L}_{cls}$ and $\mathcal{L}_{kd}$.
It should be noted that only the $C_{o2m}$ branch is deployed at inference time.

\section{Experiments}
To validate the effectiveness of \shortname{}, we conduct experiments on two widely used AU detection datasets: BP4D \cite{BP4D} and DISFA \cite{DISFA}.

\begin{table*}[h]
\centering
\resizebox{1\textwidth}{!}{
\begin{tabular}{l  |c    |c    |c    |c    |c    |c    |c    |c    |c    |c    |c    |c    |c}
\hline
Methods  &AU1    &AU2    &AU4    &AU6    &AU7    &AU10   &AU12   &AU14   &AU15   &AU17   &AU23   &AU24   &AVG\\
\hline
\hline
Baseline&51.6&39.0&60.0&72.8&79.0&79.1&87.2&62.0&49.3&58.1&48.9&51.1&61.5\\
\hline
\hline
C\_T\_$C_{o2o}$&55.8&44.9&56.9&77.8&75.6&82.8&87.5&61.3&48.7&61.9&48.2&52.5&62.8\\
C\_T\_$C_{o2m}$&52.7&47.0&56.8&75.0&75.3&82.2&88.0&63.1&51.9&64.3&49.5&53.7&63.3\\
\hline
\hline
C\_T1\_$C_{o2o}$\_T1\_$C_{o2m}$ &55.7&42.5&60.9&76.4&76.7&83.6&86.6&62.4&47.4&65.8&49.5&57.9&63.8\\
C\_T1\_$C_{o2o}$\_T2\_$C_{o2o}$&55.1&46.8&58.8&77.5&74.7&83.4&87.2&63.4&48.9&65.9&50.2&56.9&64.1\\
C\_T1\_$C_{o2m}$\_T2\_$C_{o2m}$&58.0&46.8&59.9&76.5&76.0&83.6&87.3&60.3&49.9&66.1&49.9&56.6&64.2\\
C\_T1\_$C_{o2o}$\_T2\_$C_{o2m}$ &55.4&46.0&59.8&78.7&77.7&82.7&88.6&64.7&51.4&65.7&50.9&56.0&\textbf{64.8}\\
\hline
\hline
C\_T1\_$C_{o2o}$\_T2\_$C_{o2m}$\_F&49.4&46.0&61.2&77.4&77.6&83.0&88.4&65.9&50.3&63.6&50.3&52.2&63.8\\
C\_T1\_$C_{o2o}$\_T2\_$C_{o2m}$\_R&51.8&44.8&58.7&77.6&77.4&82.5&87.6&64.7&51.7&66.0&49.7&56.2&64.1\\
C\_T1\_$C_{o2o}$\_T2\_$C_{o2m}$\_C&54.0&43.1&60.3&77.2&78.1&84.3&86.7&63.2&51.6&65.0&48.5&56.9&64.1\\
\hline
\end{tabular}}
\caption{Ablation studies on BP4D. We compare variants with various key components: with or without the transformer module, different classifiers, online knowledge distillation, and different attention drop mechanisms.}
\label{tab:ab}
\vspace{-6mm}
\end{table*}

\subsection{Implementation Details}
\noindent
\textbf{Datasets} BP4D~\cite{BP4D} contains 41 participants with 23 females and 18 males who were involved in 8 spontaneous expressions sessions. In total, 328 videos with 140,000 frames are recorded and then annotated with 12 AUs (AU1, AU2, AU4, AU6, AU7, AU10, AU12, AU14, AU15, AU17, AU23, and AU24). 
We evaluate the models with subject exclusive 3-fold cross-validation protocol following existing works \cite{JAANet,JAANet+,SRERL,TransAU}, where two folds are for training while the remaining one is for testing. 

DISFA~\cite{DISFA} involves 27 participants with 12 females and 15 males. Each person is documented in a video. The entire dataset consists of over 100,000 frames with intensity annotation ranging from 0 to 5 on 12 AUs. Following the protocol in \cite{JAANet,JAANet+,SRERL,TransAU}, we select 8 AUs (AU1, AU2, AU4, AU6, AU9, AU12, AU25, and AU26) for subject exclusive 3-fold cross-validation, and use intensity 2 as a threshold to distinguish between positive and negative samples.

\noindent
\textbf{Evaluation criteria}
We evaluate the models with F1-score \cite{jeni2013facing} that considers per AU's precision and recall and is widely used in the multi-label classification task, especially when samples are imbalanced across categories. 
Here, we calculate the F1-score on 12 AUs for BP4D and 8 AUs for DISFA. 
Per-AU's score indicates the performance of different models on individual AU while the average F1-score across all AUs exhibits one model's overall performance. 

\noindent
\textbf{Training details} 
All images are cropped based on the detection box provided by RetinaFace~\cite{retinaface}. 
We do not use facial landmarks to pre-process faces thus our input to the network is unaligned face images. 
Since we use FAN~\cite{FAN} to extract features, our network input is a 256x256 RGB face image. 
Similar to J\^AANet \cite{JAANet+}, we use random rotation (+/- 15 degrees), horizon flipping, scaling (0.75 - 1.25) and centre shifting (-10 - 10) for data augmentation. 
FAN \cite{FAN} is pre-trained and fixed while other learnable parameters are optimized using AdamW with hyper-parameters $\beta_{1}=0.9$ and $\beta_{2}=0.999$ without weight decay.
The network is trained for 12 epochs per fold with starting learning rate at 0.0001 and decaying $30\%$ every 4 epochs. 
We use the timm library \footnote{https://github.com/rwightman/pytorch-image-models} to implement our transformer with its parameter settings like MLP ratio 4 and head number = dim//64 (dim denotes feature dimension).
. 
For DISFA, following \cite{JAANet+,SRERL}, we use weights trained on BP4D as initialization and then fine-tune on DISFA. 
To calculate F1-score, we binarize the Sigmoid predictions with the threshold 0.5.
The weight balance parameter $\lambda$ is set to 0.2 after grid search. 
All the implementations are based on PyTorch \cite{paszke2017automatic}.
\begin{figure*}[t]
\centering
\subfigure[{\fontsize{7}{5}\selectfont ground-truth labels}]{\includegraphics[width=0.32\textwidth]{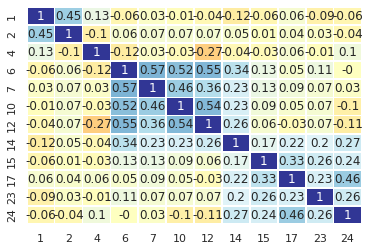}}
\subfigure[{\fontsize{7}{5}\selectfont predictions of C\_T\_$C_{o2o}$}]{\includegraphics[width=0.32\textwidth]{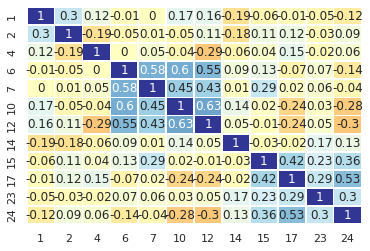}}
\subfigure[{\fontsize{7}{5}\selectfont predictions of C\_T\_$C_{o2m}$}]{\includegraphics[width=0.32\textwidth]{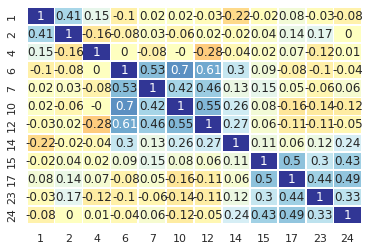}}
\caption{
Correlation maps of (a) ground-truth labels, (b) predictions of C\_T\_$C_{o2o}$ and (c)predictions of C\_T\_$C_{o2m}$, where each entry (i,j) is computed as the coefficient correlation between the i-th AU and the j-th AU.}
\label{fig:corr}
\vspace{-5mm}
\end{figure*}


\subsection{Ablation Studies}
We carry out ablation studies on BP4D to reckon the elements contributing to the efficacy of the proposed framework. 
In particular, the contributions of the transformer module, the One-to-Many classifier, the OKD instances, attention drop mechanism in the transformer are analysed. 
For fair comparison, all variants are trained with the same setting including data augmentation and training schedule. 

\noindent
\textbf{Effectiveness of Transformer}
To verify the effectiveness of the transformer module, we remove the transformer module of the proposed framework by directly performing a multi-label classification on the feature maps produced by the convolution module (see $\mathbf{F}_{c}$ in Figure~\ref{fig:overview}). 
The AU detection is done by first average pooling the 4-D dimensional tensor, then feeding the generated 1-D vector to a fully connected classifier. 
We call this variant Baseline. We implement another two variants by feeding the convolution module's output to the identical transformer module with separate classifiers $C_{o2o}, C_{o2m}$ for comparison. We call them C\_T\_$C_{o2o}$ and C\_T\_$C_{o2m}$ respectively. At this point, no knowledge distillation part is included.

The results are in Table \ref{tab:ab}.
Compared with Baseline, both C\_T\_$C_{o2o}$ and C\_T\_$C_{o2m}$ get higher F1-score in the majority of AUs with obvious improvement on AU2, AU6, AU17, AU23.
Overall, in terms of the average F1-score, C\_T\_$C_{o2o}$ and C\_T\_$C_{o2m}$ exceeds Baseline by 1.3\% and 1.8\% separately. 
We believe the advancement comes from the more representative features learned by the transformer module. 
Besides, simple Baseline alone surpasses JAANet\cite{JAANet}, DSIN \cite{DSIN} with more complex feature extractors, which reveals the strong representation ability of FAN features.  

\noindent
\textbf{Effectiveness of classifiers}
C\_T\_$C_{o2m}$ outperforms C\_T\_$C_{o2o}$ by 0.5\% in average F1-score. We argue C\_T\_$C_{o2m}$ is superior to $C_{o2o}$ because the former one considers the AU co-occurrences in both feature learning and classification stages while the latter only captures this characteristic in feature learning within the transformer module. 
This argument is evidenced by Figure \ref{fig:corr} which presents the correlation coefficients between pairwise AUs from annotations, predictions of $C_{o2o}$ and predictions of $C_{o2m}$ separately. 
For example, AU14 is associated with AU6, AU7, AU10, AU12 in Figure~\ref{fig:corr}(a). 
This pattern is captured by C\_T\_$C_{o2m}$ but ignored in C\_T\_$C_{o2o}$.
We assume that the $C_{o2m}$ is sufficient in detecting the lower part of AUs because AU14 fosters inter connections. 
In addition, we compute the element-wise Euclidean distance between correlation matrices for more direct comparison. 
The distance between labels and predictions from $C_{o2o}$ is 0.012 while it decreases to 0.008 for $C_{o2m}$, which further confirms the ability of $C_{o2m}$ in learning AU associations from label statistics. 

\noindent
\textbf{Improvement from OKD training}
We first test a variant C\_T1\_$C_{o2o}$\_T1\_$C_{o2m}$ which shares the feature extraction (C\_T1) but is fed into two classifiers $C_{o2o}$ and $C_{o2m}$. 
From the table, this variant gains performance over student alone (\ie C\_T\_$C_{o2o}$ and C\_T\_$C_{o2m}$). 

Such improvement motivates us to explore key factors concerning the OKD: where to put split point and how to increase peers' diversity. 
We set up three variants: 
\one~\Vo{}, \two~\Vm{}, \three~\Vom{}.
The architecture for \three{} is in Figure \ref{fig:overview}, in which the breakpoint for two branches is after the convolution module.
\one{} replaces $C_{o2m}$ in T2 branch of \three{} with $C_{o2o}$ while \two{} replaces $C_{o2o}$ in T1 branch of \three{} with $C_{o2m}$.
From the Table \ref{tab:ab}, all the variants obtain the average F1-score gains with \three{} ranking 1st.
\one{} and \two{} outperform C\_T1\_$C_{o2o}$\_T1\_$C_{o2m}$  by 0.3\% and 0.4\% respectively because the same feature descriptor hurts the diversity.
This finding is supported by the analysis of branch diversity proposed in \cite{li2021online,chen2020online}. 

From the results, we observe that
\one{} improves C\_T1\_$C_{o2o}$ from 62.8\% to 64.1\% while \two{} improves C\_T1\_$C_{o2m}$ from 63.3\% to 64.2\%. 
Both of them are comparable to the SOTA performance (see Table \ref{tab:bp4d_sota}).
It is worth mentioning here that this is the \textit{first} time that OKD is deployed on facial AU detection task, and it obtains significant performance gains. 

\three{} achieves the best performance. 
More concretely, the output of $C_{o2o}$ in \three{} gets 64.7\% in average F1-score, the output of $C_{o2m}$ in \three{} achieves 64.8\% as well as $P_t$.
The superior performance shows that designing two branches with different classifiers boosts the peer diversity, which is effective in OKD \cite{chen2020online}.
We set default branch as 2 because we leveraged two different classifiers.
We have tried increasing the branch to three which obtains another 0.1\% gain.


\noindent
\textbf{Effectiveness of attention drop in Transformer}
In the following experiment, we dig into how the attention map influences the AU features.
We consider four variants: 
(a) \Vom{}\_F, 
(b) \Vom{}\_R, 
(c) \Vom{}\_C 
and (d) \Vom{}. 
The description is in Figure \ref{fig:tranformer}.
In (a), a full attention map is used, which means all elements in $\mathbf{A}$ will be multiplied with $\mathbf{V}$ to produce new AU features.
(b) deactivates the lower similarity in each row of $\mathbf{A}$.
(c) deactivates the lower similarity in each column of $\mathbf{A}$.
(d) learns a binary mask. 
As is shown in Table \ref{tab:ab}, 
(a), (b) and (c) improve the average F1-score of C\_T\_$C_{o2m}$ from 63.3\% to 63.8\%, 61.4\% and 64.1\% respectively. 
It demonstrates that all attention mechanisms benefit the OKD framework. 
(b) and (c) outperforming (a) may be caused by the alleviation of over-fitting in the full attention map. 
The learnable attention drop stands out by increasing the F1-score in C\_T\_$C_{o2m}$ from 63.3\% to 64.8\%. 
\begin{figure}[t]
\vspace{-2mm}
\centering
\includegraphics[width=0.38\textwidth]{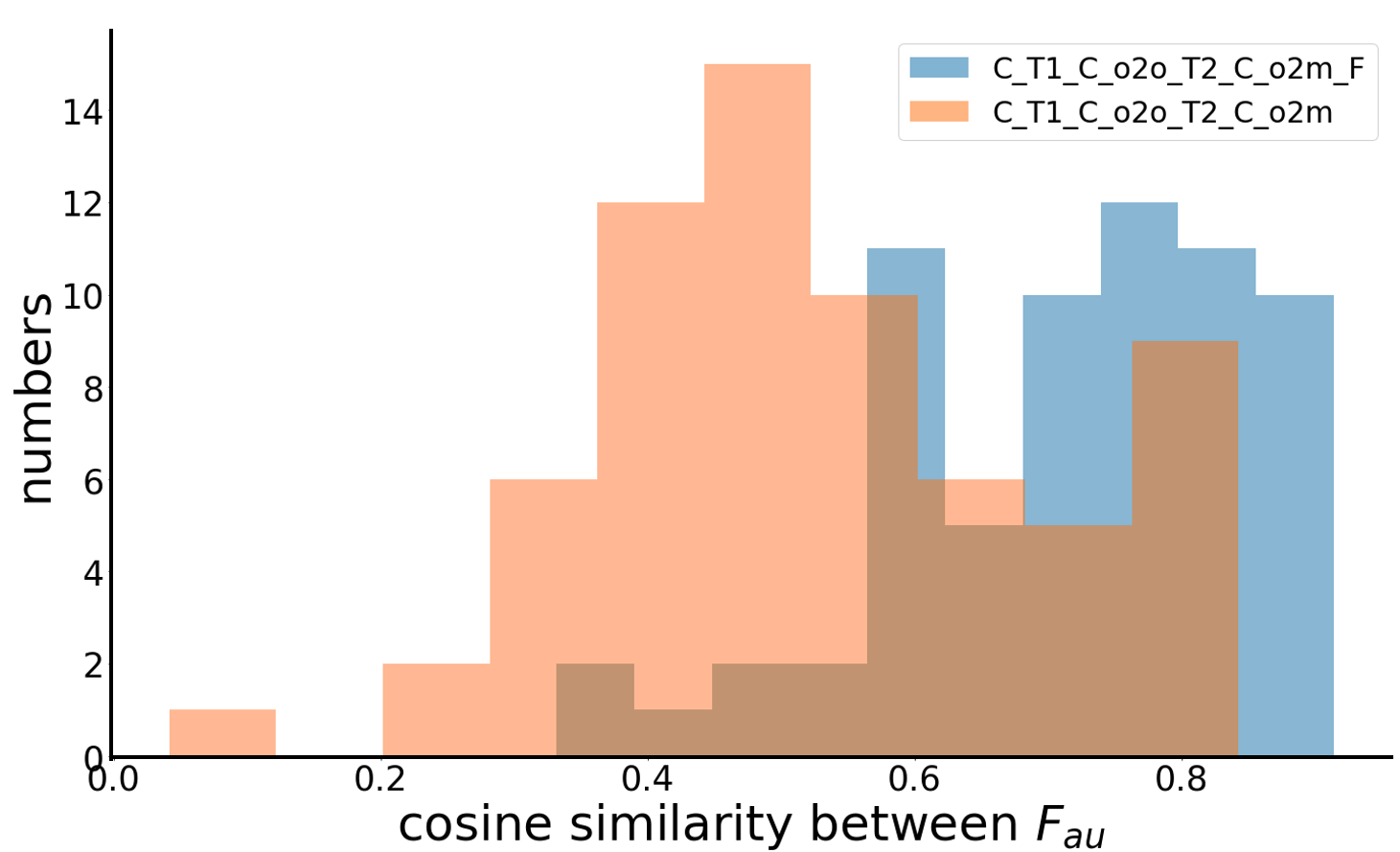}
\caption{The pairwise cosine similarity distribution among AU features $\mathbf{F}_{au}$ before $C_{o2m}$ for variant (a) and (d).}
\vspace{-4mm}
\label{fig:fau_distribution}
\end{figure}

Figure \ref{fig:fau_distribution} illustrates the distribution of the pairwise cosine distance among $\mathbf{F}_{au}$. 
By comparison, the average similarity is decreased from 0.738~(a) to 0.566~(d).
It shows attention drop will increase diversity among different AU representations.

\begin{table}[h]
\parbox{.48\textwidth}{
\centering
\begin{tabular}{c|c|c|c|c}
\hline
Ms & BL & BL-FAN+R18 & BL-LF &BL-HM \\
\hline
F1-score & 61.5 &59.5 &59.2 &60.8\\
\hline
\end{tabular}
\caption{Impact of FAN feature on BP4D (F1-score in \%). \label{tab:ab_fea}}}
\hfill

\parbox{.48\textwidth}{
\centering
\begin{tabular}{c|c|c|c|c|c}
\hline
Encode No. &1&3 &5 &7 &9 \\
\hline
F1-score&63.8&64.4&64.8&\textbf{64.9}&64.1\\
\hline
\end{tabular}
\caption{Impact of No. of TransBlock on BP4D (F1-score in \%). \label{tab:encode}}}
\hfill

\parbox{.48\textwidth}{
\centering
\begin{tabular}{c|c|c|c|c|c}
\hline
$\lambda$ &0.05 &0.1 &0.2 &0.5 &1  \\
\hline
F1-score&64.1&64.3&\textbf{64.8}&63.9&63.8\\
\hline
\end{tabular}}
\caption{Sensitivity to $\lambda$ for FAN-Trans on BP4D (F1-score in \%). \label{tab:hyper}}
\vspace{-10mm}
\end{table}

\begin{table*}[t]
\centering
\resizebox{1\textwidth}{!}{
\begin{tabular}{l  |c    |c    |c    |c    |c    |c    |c    |c    |c    |c    |c    |c    |c}
    \hline
Methods  &AU1    &AU2    &AU4    &AU6    &AU7    &AU10   &AU12   &AU14   &AU15   &AU17   &AU23   &AU24   &AVG\\
\hline
\hline
JPML~\cite{JPML}    &32.6&25.6&37.4&42.3&50.5&72.2&74.1&65.7&38.1&40.0&30.4&42.3&45.9\\
DRML~\cite{DRML}    &36.4&41.8&43.0&55.0&67.0&66.3&65.8&54.1&33.2&48.0&31.7&30.0&48.3\\
JAANet~\cite{JAANet}  &47.2&44.0&54.9&77.5&74.6&[84.0]&86.9&61.9&43.6&60.3&42.7&41.9&60.0\\
J\^AANet~\cite{JAANet+}&53.8&47.8&58.2&[78.5]&75.8&82.7&88.2&63.7&43.3&61.8&45.6&49.9&62.4\\
DSIN~\cite{DSIN}    &51.7&40.4&56.0&76.1&73.5&79.9&85.4&62.7&37.3&62.9&38.8&41.6&58.9\\
SRERL~\cite{SRERL}   &46.9&45.3&55.6&77.1&[78.4]&83.5&[87.6]&63.9&[52.2]&[63.9]&47.1&53.3&62.1\\
UGN-B~\cite{UGN-B}   &[54.2]&[46.4]&56.8&76.2&76.7&82.4&86.1&64.7&51.2&63.1&48.6&53.6&63.3\\
TransAU~\cite{TransAU} &51.7&\textbf{49.3}&61.0&77.8&\textbf{79.5}&82.9&86.3&\textbf{67.6}&51.9&63.0&43.7&\textbf{56.3}&64.2\\
MONET~\cite{tallec2022multi}&54.5 &45.0 &\textbf{61.5} &75.9 &78.0 &\textbf{84.5} &87.6 &[65.1] &\textbf{54.8} &60.5 &\textbf{53.0} &53.2 &64.5\\
\hline
\shortname{}&~\textbf{55.4}&46.0&[59.8]&\textbf{78.7}&77.7&82.7&\textbf{88.6}&64.7&51.4&\textbf{65.7}&[50.9]&[56.0]&\textbf{64.8}\\
\hline
\end{tabular}}
\caption{Comparison of the classification results (F1-score in \%) with other methods on BP4D. Bold numbers indicate the best performance; Bracketed numbers indicate the second best.}
\label{tab:bp4d_sota}
\end{table*}

\begin{table*}[t]
\centering
\begin{tabular}{l        |c    |c    |c    |c    |c    |c    |c    |c   |c}
\hline
Methods  &AU1    &AU2    &AU4    &AU6    &AU9    &AU12  &AU25   &AU26 &AVG\\
\hline
\hline
DRML~\cite{DRML}&17.3&17.7&37.4&29.0&10.7&37.7&38.5&20.1&26.7\\
JAANet~\cite{JAANet}  &43.7&46.2&56.0&41.4&44.7&69.6&88.3&58.4&56.0\\
J\^AANet~\cite{JAANet+}&\textbf{62.4}&\textbf{60.7}&67.1&41.1&45.1&73.5&90.9&\textbf{67.4}&63.5\\
DSIN~\cite{DSIN}&42.4&39.0&68.4&28.6&46.8&70.8&90.4&42.2&53.6\\
SRERL~\cite{SRERL}   &45.7&47.8&59.6&47.1&45.6&73.5&84.3&43.6&55.9\\
UGN-B~\cite{UGN-B}   &43.3&48.1&63.4&49.5&48.2&71.9&90.8&59.0&60.0\\ 
TransAU~\cite{TransAU}&46.1&48.6&\textbf{72.8}&\textbf{56.7}&[50.0]&72.1&90.8&55.4&61.5\\
MONET~\cite{tallec2022multi}& 55.8 &[60.4] &68.1 &[49.8] &48.0 &[73.7] &[92.3] &[63.1] &\textbf{63.9}\\
\hline
\shortname{}&[56.4]&50.2&[68.6]&49.2&\textbf{57.6}&\textbf{75.6}&\textbf{93.6}&58.8&[63.8]\\
\hline
\end{tabular}
\caption{Comparison of the classification results (F1-score in \%) with other methods on DISFA. Bold numbers indicate the best performance; Bracketed numbers indicate the second best.}
\label{tab:disfa_sota}
\vspace{-4mm}
\end{table*}
\vspace{-2mm}

\noindent
\textbf{Impact of FAN feature}
We compare four variants in Table \ref{tab:ab_fea}: BL is the baseline method; BL-FAN+R18 replaces FAN in BL with feature map of same resolution (64x64) from ResNet18 (after 2rd layer) pretrained on ImageNet; BL-LF is BL without LF; BL-HW is BL without the product of heatmap.
Based on results, we conclude features pretrained on the alignment task are superior to generic features from a classification model.
Additionally, although HM is inferior to LF for overall contribution, it complements LF by giving more attention around landmarks which are recognized as AU active areas.

\noindent
\textbf{Encode numbers in transformer module}
We set encode number in the transformer module as 5 in the main experiment to keep balance with convolution operations.
Table \ref{tab:encode} presents the performance influenced by the transBlock number in the transformer module on the BP4D.
We test 5 cases with encode numbers as 1,3,5,7,9 respectively.
From this table, we can see that if very few encodes are utilized, the performance will drop quickly.
This is caused by the limited learning capacity in the transformer module. 
Besides, when the number of encoding increases to 9, the performance will also drop because of too much flexibility within the two branches.
Although encode number 7 gets a slightly better F1-score than encode number 5, considering the increased model complexity, we still use encode number 5 in our final model. 

\noindent
\textbf{Sensitivity to hyper-parameter} $\boldsymbol{\lambda}$ 
Basically, we test 5 weights for $\lambda$: 0.05, 0.1, 0.2, 0.5 and 1.
Table~\ref{tab:hyper} illustrates the impact of hyper-parameter $\lambda$ in the proposed framework. 
From these results, we can see that our method is affected by the weights to balance the distillation loss and the classification loss. 
The too small or too large value will deteriorate the performance gains. 
Thus, a grid search technique contributes to the best model. 

Finally, the best configuration \Vom{} by setting $\lambda$ as 0.2 is \shortname{}.

\subsection{Comparison with State-of-the-Art methods}
We compare \shortname{} with published AU detection techniques including the methods focusing on attention or regional feature \ie JPML~\cite{JPML}, DRML~\cite{DRML}, JAANet~\cite{JAANet}, J\^AANet~\cite{JAANet+}, methods taking the relationship of AUs into account \ie DSIN~\cite{DSIN}, SRERL~\cite{SRERL}, UGN-B~\cite{UGN-B} and very lately proposed works TransAU~\cite{TransAU} and MONET~\cite{tallec2022multi}.
The results for other methods are taken from the papers  \cite{TransAU,PIAP}.
Table~\ref{tab:bp4d_sota} shows the performance comparison on the BP4D~\cite{BP4D}. 
The proposed method performs better than the SOTA methods with an average F1-score 64.8\%.

Table~\ref{tab:disfa_sota} compares the performance of our proposed \shortname{} with SOTA methods on the DISFA~\cite{DISFA}. 
It can be seen that our method obtains a 63.8\% average F1-score. 
It is 2\% better than TransAU \cite{TransAU} which also deploys transformer in model design. 

\vspace{-2mm}
\section{Conclusion and Discussion}
\vspace{-2mm}
\noindent
\textbf{Conclusion}
In this work, we propose \shortname{} for facial AU detection which can learn representative AU features and correlations among both AU features and AU labels in an OKD framework. \shortname{} takes advantage of the multi-scale face alignment feature maps to learn AU features from AU active regions with heatmap attention. 
It uses transformer to model AU co-occurrences with a learnable binary mask to drop self attention in order to discriminate different AU features.
It customizes OKD with diverse classifiers designed for AU detection.
Experiments show its advantages over SOTA methods.

\noindent
\textbf{Discussion}
\shortname{} is built on a pre-trained face alignment model for extraction of AU features. Thus the performance of face alignment will influence the model performance. 
Besides, as is show in Figure \ref{fig:attention}, although the face alignment features provide a strong representation and the attention area is roughly around key points, it is still hard to to assign them very precisely without manual supervision.
Exploring how to learn individual AU attentions associated with landmarks is a promising direction.
\clearpage
{\small
\bibliographystyle{ieee_fullname}
\bibliography{egbib}

\begin{thebibliography}{10}\itemsep=-1pt

\bibitem{FAN}
Adrian Bulat and Georgios Tzimiropoulos.
\newblock How far are we from solving the 2d \& 3d face alignment problem?(and
  a dataset of 230,000 3d facial landmarks).
\newblock In {\em ICCV}, 2017.

\bibitem{chen2020online}
Defang Chen, Jian-Ping Mei, Can Wang, Yan Feng, and Chun Chen.
\newblock Online knowledge distillation with diverse peers.
\newblock In {\em AAAI}, 2020.

\bibitem{DSIN}
Ciprian Corneanu, Meysam Madadi, and Sergio Escalera.
\newblock Deep structure inference network for facial action unit recognition.
\newblock In {\em ECCV}, 2018.

\bibitem{retinaface}
Jiankang Deng, Jia Guo, Evangelos Ververas, Irene Kotsia, and Stefanos
  Zafeiriou.
\newblock Retinaface: Single-shot multi-level face localisation in the wild.
\newblock In {\em CVPR}, 2020.

\bibitem{dosovitskiy2020image}
Alexey Dosovitskiy, Lucas Beyer, Alexander Kolesnikov, Dirk Weissenborn,
  Xiaohua Zhai, Thomas Unterthiner, Mostafa Dehghani, Matthias Minderer, Georg
  Heigold, Sylvain Gelly, et~al.
\newblock An image is worth 16x16 words: Transformers for image recognition at
  scale.
\newblock In {\em ICLR}, 2020.

\bibitem{FACS}
Paul Ekman and Erika~L Rosenberg.
\newblock {\em What the face reveals: Basic and applied studies of spontaneous
  expression using the Facial Action Coding System (FACS)}.
\newblock Oxford University Press, USA, 1997.

\bibitem{KD}
Geoffrey Hinton, Oriol Vinyals, and Jeff Dean.
\newblock Distilling the knowledge in a neural network.
\newblock {\em arXiv}, 2015.

\bibitem{TransAU}
Geethu~Miriam Jacob and Bjorn Stenger.
\newblock Facial action unit detection with transformers.
\newblock In {\em CVPR}, 2021.

\bibitem{jeni2013facing}
L{\'a}szl{\'o}~A Jeni, Jeffrey~F Cohn, and Fernando De~La~Torre.
\newblock Facing imbalanced data--recommendations for the use of performance
  metrics.
\newblock In {\em 2013 Humaine association conference on affective computing
  and intelligent interaction}, 2013.

\bibitem{lan2018knowledge}
Xu Lan, Xiatian Zhu, and Shaogang Gong.
\newblock Knowledge distillation by on-the-fly native ensemble.
\newblock In {\em NeurIPS}, 2018.

\bibitem{SRERL}
Guanbin Li, Xin Zhu, Yirui Zeng, Qing Wang, and Liang Lin.
\newblock Semantic relationships guided representation learning for facial
  action unit recognition.
\newblock In {\em AAAI}, 2019.

\bibitem{li2021online}
Zheng Li, Jingwen Ye, Mingli Song, Ying Huang, and Zhigeng Pan.
\newblock Online knowledge distillation for efficient pose estimation.
\newblock In {\em ICCV}, 2021.

\bibitem{lin2021end}
Kevin Lin, Lijuan Wang, and Zicheng Liu.
\newblock End-to-end human pose and mesh reconstruction with transformers.
\newblock In {\em CVPR}, 2021.

\bibitem{DISFA}
S~Mohammad Mavadati, Mohammad~H Mahoor, Kevin Bartlett, Philip Trinh, and
  Jeffrey~F Cohn.
\newblock Disfa: A spontaneous facial action intensity database.
\newblock {\em TAC}, 2013.

\bibitem{hourglass}
Alejandro Newell, Kaiyu Yang, and Jia Deng.
\newblock Stacked hourglass networks for human pose estimation.
\newblock In {\em ECCV}, 2016.

\bibitem{paszke2017automatic}
Adam Paszke, Sam Gross, Soumith Chintala, Gregory Chanan, Edward Yang, Zachary
  DeVito, Zeming Lin, Alban Desmaison, Luca Antiga, and Adam Lerer.
\newblock Automatic differentiation in pytorch.
\newblock 2017.

\bibitem{JAANet}
Zhiwen Shao, Zhilei Liu, Jianfei Cai, and Lizhuang Ma.
\newblock Deep adaptive attention for joint facial action unit detection and
  face alignment.
\newblock In {\em ECCV}, 2018.

\bibitem{JAANet+}
Zhiwen Shao, Zhilei Liu, Jianfei Cai, and Lizhuang Ma.
\newblock Jaa-net: joint facial action unit detection and face alignment via
  adaptive attention.
\newblock {\em IJCV}, 2021.

\bibitem{UGN-B}
Tengfei Song, Lisha Chen, Wenming Zheng, and Qiang Ji.
\newblock Uncertain graph neural networks for facial action unit detection.
\newblock In {\em AAAI}, 2021.

\bibitem{song2021facial}
Wenyu Song, Shuze Shi, and Gaoyun An.
\newblock Facial action unit detection based on transformer and attention
  mechanism.
\newblock In {\em ICIG}, 2021.

\bibitem{tallec2022multi}
Gauthier Tallec, Arnaud Dapogny, and Kevin Bailly.
\newblock Multi-order networks for action unit detection.
\newblock {\em TAC}, 2022.

\bibitem{PIAP}
Yang Tang, Wangding Zeng, Dafei Zhao, and Honggang Zhang.
\newblock Piap-df: Pixel-interested and anti person-specific facial action unit
  detection net with discrete feedback learning.
\newblock In {\em ICCV}, 2021.

\bibitem{emonet}
Antoine Toisoul, Jean Kossaifi, Adrian Bulat, Georgios Tzimiropoulos, and Maja
  Pantic.
\newblock Estimation of continuous valence and arousal levels from faces in
  naturalistic conditions.
\newblock {\em Nature Machine Intelligence}, 2021.

\bibitem{vaswani2017attention}
Ashish Vaswani, Noam Shazeer, Niki Parmar, Jakob Uszkoreit, Llion Jones,
  Aidan~N Gomez, {\L}ukasz Kaiser, and Illia Polosukhin.
\newblock Attention is all you need.
\newblock In {\em NeurIPS}, 2017.

\bibitem{walecki2017deep}
Robert Walecki, Vladimir Pavlovic, Bj{\"o}rn Schuller, Maja Pantic, et~al.
\newblock Deep structured learning for facial action unit intensity estimation.
\newblock In {\em CVPR}, 2017.

\bibitem{score-cam}
Haofan Wang, Zifan Wang, Mengnan Du, Fan Yang, Zijian Zhang, Sirui Ding, Piotr
  Mardziel, and Xia Hu.
\newblock Score-cam: Score-weighted visual explanations for convolutional
  neural networks.
\newblock In {\em CVPRW}, 2020.

\bibitem{wang2013capturing}
Ziheng Wang, Yongqiang Li, Shangfei Wang, and Qiang Ji.
\newblock Capturing global semantic relationships for facial action unit
  recognition.
\newblock In {\em ICCV}, 2013.

\bibitem{FANFace}
Jing Yang, Adrian Bulat, and Georgios Tzimiropoulos.
\newblock Fan-face: a simple orthogonal improvement to deep face recognition.
\newblock In {\em AAAI}, 2020.

\bibitem{yang2021transpose}
Sen Yang, Zhibin Quan, Mu Nie, and Wankou Yang.
\newblock Transpose: Keypoint localization via transformer.
\newblock In {\em ICCV}, 2021.

\bibitem{zhang2019your}
Linfeng Zhang, Jiebo Song, Anni Gao, Jingwei Chen, Chenglong Bao, and Kaisheng
  Ma.
\newblock Be your own teacher: Improve the performance of convolutional neural
  networks via self distillation.
\newblock In {\em ICCV}, 2019.

\bibitem{zhang2019scan}
Linfeng Zhang, Zhanhong Tan, Jiebo Song, Jingwei Chen, Chenglong Bao, and
  Kaisheng Ma.
\newblock Scan: A scalable neural networks framework towards compact and
  efficient models.
\newblock {\em NeurIPS}, 2019.

\bibitem{BP4D}
Xing Zhang, Lijun Yin, Jeffrey~F Cohn, Shaun Canavan, Michael Reale, Andy
  Horowitz, Peng Liu, and Jeffrey~M Girard.
\newblock Bp4d-spontaneous: a high-resolution spontaneous 3d dynamic facial
  expression database.
\newblock {\em IVC}, 2014.

\bibitem{zhang2018classifier}
Yong Zhang, Weiming Dong, Bao-Gang Hu, and Qiang Ji.
\newblock Classifier learning with prior probabilities for facial action unit
  recognition.
\newblock In {\em CVPR}, 2018.

\bibitem{JPML}
Kaili Zhao, Wen-Sheng Chu, Fernando De~la Torre, Jeffrey~F Cohn, and Honggang
  Zhang.
\newblock Joint patch and multi-label learning for facial action unit
  detection.
\newblock In {\em CVPR}, 2015.

\bibitem{DRML}
Kaili Zhao, Wen-Sheng Chu, and Honggang Zhang.
\newblock Deep region and multi-label learning for facial action unit
  detection.
\newblock In {\em CVPR}, 2016.

\end{thebibliography}
}
\end{document}